\documentclass[journal]{IEEEtran}
\usepackage{xcolor,soul,framed} %,caption
\colorlet{shadecolor}{yellow}
\usepackage[pdftex]{graphicx}
\graphicspath{{../pdf/}{../jpeg/}}
\DeclareGraphicsExtensions{.pdf,.jpeg,.png}
\usepackage[cmex10]{amsmath}
\usepackage{array}
\usepackage{mdwmath}
\usepackage{mdwtab}
\usepackage{eqparbox}
\usepackage{url}
\usepackage[export]{adjustbox}
\usepackage{subcaption}
\usepackage{atbegshi}% http://ctan.org/pkg/atbegshi
\AtBeginDocument{\AtBeginShipoutNext{\AtBeginShipoutDiscard}}
\begin{document}
\bstctlcite{IEEEexample:BSTcontrol}
    \title{AI Technical Considerations:\\
 Data Storage, Cloud usage and AI Pipeline}
  \author{P.M.A van Ooijen$^{1}$,
      ~Erfan Darzidehkalani$^{1}$,  Andre Dekker$^{2}$ \\
     \vspace{10pt}$^{1}$Machine Learning Lab, Data Science Center in Health (DASH),University Medical Center Groningen,University of Groningen, Hanzeplein 1, The Netherlands. \\
    $^{2}$Department of Radiation Oncology (MAASTRO)GROW School for Oncology and Developmental Biology, Maastricht University Medical Centre+, Maastricht,The Netherlands.\\
     p.m.a.van.ooijen@umcg.nl, e.darzidehkalani@umcg.nl, andre.dekker@maastro.nl}  
    %   Tibault~Reveyrand,~\IEEEmembership{Member,~IEEE,}\\
    %   Ignacio~Ramos,~\IEEEmembership{Student Member,~IEEE,}
    %   Erez Falkenstein,~\IEEEmembership{Student Member,~IEEE,}
    %   and~Zoya~Popovi\'c,~\IEEEmembership{Fellow,~IEEE}% <-this % stops a space
%   \thanks{Manuscript received July 10, 2012. \hl{This paper is an expanded paper from the IEEE MTT-S Int. Microwave Symposium held on June 17-22, 2012, in Montreal, Canada.} This work was funded in part by the Office of Naval Research under the Defense Advanced Research Projects Agency (DARPA) Microscale Power Conversion (MPC) Program under Grant N00014-11-1-0931, and in part by the Advanced Research Projects Agency-Energy (ARPA-E), U.S. Department of Energy, under Award Number DE-AR0000216.}
%   \thanks{M. Roberg is with TriQuint Semiconductor, 500 West Renner Road Richardson, TX 75080 USA (e-mail: michael.roberg@tqs.com).}% <-this % stops a space
%   \thanks{T. Reveyrand is with the XLIM Laboratory, UMR 7252, University of Limoges, 87060 Limoges, France (e-mail: tibault.reveyrand@xlim.fr).}%
%   \thanks{I. Ramos and Z. Popovic are with the Department of Electrical, Computer and Energy Engineering, the University of Colorado, Boulder, CO, 80309-0425 USA (e-mail: ignacio.ramos@colorado.edu; zoya.popovic@colorado.edu).}% <-this % stops a space
\thanks{{ }} % 
% \markboth{IEEE TRANSACTIONS ON MICROWAVE THEORY AND TECHNIQUES, VOL.~60, NO.~12, DECEMBER~2012
% }{Roberg \MakeLowercase{\textit{et al.}}: High-Efficiency Diode and transistor Rectifiers}
% ====================================================================
\maketitle
% === ABSTRACT ====================================================================
% =================================================================================
\begin{abstract}
Artificial intelligence (AI), especially deep learning, requires vast amounts of data for training, testing, and validation. Collecting these data and the corresponding annotations requires the implementation of imaging biobanks that provide access to these data in a standardized way. This requires careful design and implementation based on the current standards and guidelines and complying with the current legal restrictions. However, the realization of proper imaging data collections is not sufficient to train, validate and deploy AI as resource demands are high and require a careful hybrid implementation of AI pipelines both on-premise and in the cloud.
This chapter aims to help the reader when technical considerations have to be made about the AI environment by providing a technical background of different concepts and implementation aspects involved in data storage, cloud usage, and AI pipelines.
\end{abstract}

% keywords can be removed

\IEEEpeerreviewmaketitle

\section{Introduction}
It is a well-known fact that Artificial Intelligence (AI) performance depends heavily on the availability of data and the corresponding annotations or labels that define the ground truth about the (clinical) question at hand. To obtain this data, they need to be made available by multiple institutions geographically distributed worldwide. One option is to centralize data, which requires de-identification and informed consent that includes the possibility of wide distribution and sharing of the acquired data [1]. A second option is to apply federated learning principles in which data does not have to move but does require wide distribution of the AI application and computing resources [2]. In both options, gathering extensive collections of imaging data of the same pathology or disease and a pre-defined population is a challenging task, and many projects suffer from the limitations of a small dataset. Not only because of the limited availability of the data but also because the willingness of healthcare institutes to share medical data is low. 
To tackle this problem and enable the collection of large imaging databases, careful technical considerations are required concerning the IT infrastructure. Three essential parts of this IT infrastructure that can help drive AI forward are data storage, cloud usage, and AI pipeline implementation. This chapter will dive into these three more prominent topics by addressing more specific topics such as imaging biobanking, cloud storage, and computing, federated learning, and different implementation versions of the AI pipeline in medical imaging.

\section{Data storage}

In 2010 the Quantitative Imaging Network (QIN) performed a questionnaire among their members to explore informatics methods in imaging research [3]. One of their goals was to share data among institutes to accelerate quantitative imaging research. Significant findings were a considerable variation in tools used for the de-identification, local image file storage was varying from XNAT and (commercial) Picture Archiving and Communication System (PACS) solutions to open consumer platforms such as Dropbox and local image meta-data databases varied from dedicated tools like RedCAP to simple spreadsheets on a local hard drive or USB stick.

\subsection{Imaging Biobanking}
According to the ESR position paper on imaging biobanks, a biobank is an “Organised database of medical images and associated imaging biomarkers (radiology and beyond) shared among multiple researchers and linked to other biorepositories” [4]. The key points resulting from this work were:
\begin{itemize}
 \item Imaging biobanks are "shared databases of imaging biomarkers, linked to biorepositories”
    \item Exploitation of traditional and imaging biobanks is meaningful for “personalised medicine”
    \item A European imaging biobank network would significantly boost research in the imaging domain
\end{itemize}
   
The immediate purpose of an imaging biobank is to allow the generation of imaging biomarkers for use in research studies (either using ‘conventional’ techniques or AI/Deep Learning) and to support biological validation of existing and novel imaging biomarkers. The long-term scope of imaging biobanks is the creation of a network/federation of such repositories.
An imaging biobank can exist in different scenarios. First, an imaging biobank could contain clinical research data gathered in clinical research/trials. Second, it could contain disease-specific data from clinical practice or screening programs (e.g., breast, lung, and colon cancer) based on disease characteristics. These data collections are not necessarily directly connected to a clinical research question. Third and final, an imaging biobank can contain general population data. In this case, data is collected from the general population (not only patients or population at risk), with no specific goal or disease-oriented approach. This usually involves the collection of long-term longitudinal data.

Imaging biobanks can exist as a single entity with centralized data collection or storage, but federations of biobanks are also possible to exchange data or research questions. Such a federated setup of imaging biobanks requires a centralized database or catalog of data collections present at different local imaging biobanks [4], preferably using Findable Accessible Interoperable and Reusable (FAIR) data principles [5]. Utilization of the data in such a federated environment could involve collecting the imaging data at the site where they are needed when needed, but also to have the data remain at the site of acquisition or collection and allowing software tools to access that data to perform analysis with only returning the analysis results [6].

A standard FAIR data model is required to allow the merging of (imaging) data collected from multiple biobanks. This shared data model should include typical data schemes, standard nomenclatures, references to common ontologies, etc. Furthermore, access policies should be implemented to request and grant access to data collections for specific research projects.\\
The ESR Position paper on Imaging Biobanks [4] lists several requirements for the appropriate implementation and use of imaging biobanks:
The aim should be to achieve an eco-system with a central and federated approach where query, analysis, and retrieval across multiple data repositories should be possible.
Data access should be secure and permission-based.
Proper de-identification of both imaging and meta-data should be in place.
Standard information models and terminologies should be used.

The Quantitative Imaging Network proposed a system architecture to enable data sharing, collaborative experimentation, and translation of methods to clinical practice [3].  This system architecture proposes a four-layer design. It includes informatics tools to support data repositories in a data storage layer (in blue) based on standard information models and shared semantics (in red).

The data repositories in the storage layer can be accessed by different processing and analysis in the Research Methods layer. Nowadays, this layer is extended by the training, validation, and deployment of AI systems. Finally, the clinical systems layer provides the clinical users with the tools to access the data in the underlying layers, including the results from the algorithms in the research methods layer.
Although this model was proposed in 2012, it still holds and provides a clear picture of the information used in the imaging domain.
\begin{figure}[h!]
 \centering
  \includegraphics[scale=0.42]{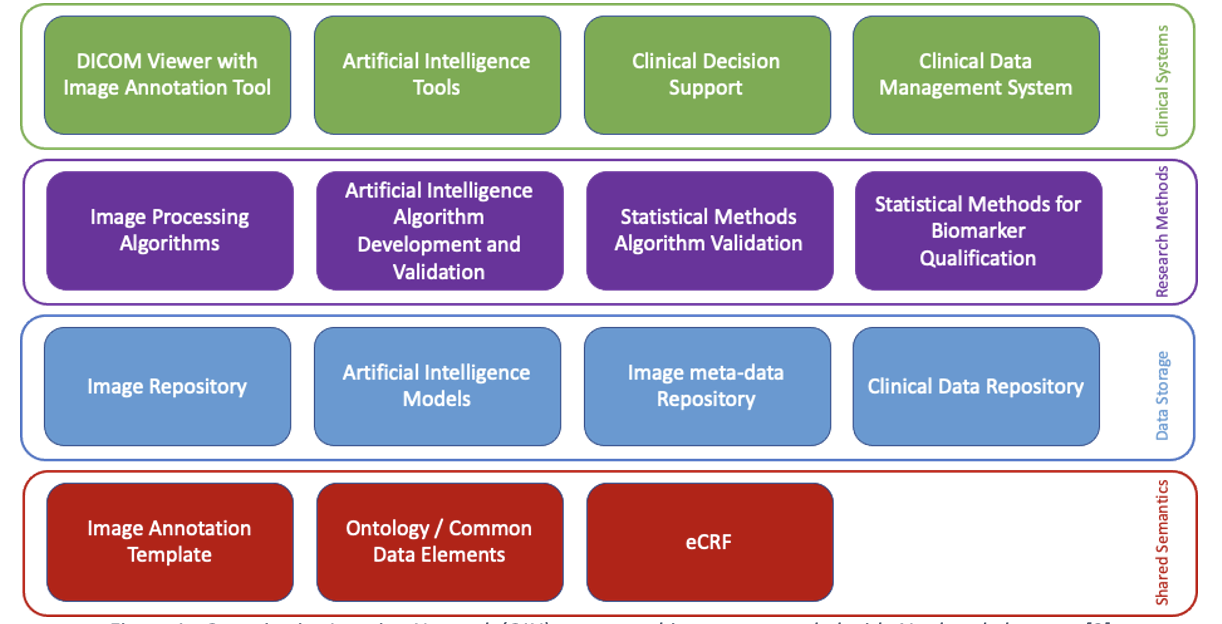}
    \caption{Quantitative Imaging Network (QIN) system architecture expanded with AI related elements}

  \label{fig:CDS}
\end{figure}

Quite a large body of work exists on the design and implementation of data environments for medical imaging, both for research and clinical practice [6] [7] [8]. Commonly used open-source tools are XNAT (imaging) and RedCap (metadata), but also other environments and implementations, both commercial and open-source, [6] [7], exist. One common denominator of the environments is that they can all operate in the ‘cloud’ of the world wide web. Known advantages of cloud usage are economy of scale, improved performance, data portability, increased and flexible storage capacity, data migration, and patient-centric connected systems [9].

\subsection{Data storage:  Challenges and Considerations}

In order for AI pipelines to work properly, data must be annotated which introduces the challenge of obtaining a proper ground truth. Ground-truth images in imaging biobanks are discrepant most of the times as medical experts are not fully concordant in their annotations. \\
Also, the quality of the data collection is vital, and several aspects have to be considered when collecting and storing data.  In medical imagining domain, numerous scanners, imaging protocols and parameter choices exist, which results in images of significantly different distributions. It is important to store the provenance of the images, i.e. by which protocol, scanners, parameters they were obtained, whether they contain missing slices or artefacts etc. It enables AI algorithms to decide what data to query and to avoid improper samples. As an example, motion corrupted MR images cause substantial problems in AI algorithms for diagnosis or segmentation [10]; whereas they are useful for motion-correction AI models. 

\section{Cloud usage}

The National Institute of Standards and Technology (NIST) defines cloud computing as follows

\textit{“Cloud computing is a model for enabling ubiquitous, convenient, on-demand network access to a shared pool of configurable computing resources (e.g., networks, servers, storage, applications, and services) that can be rapidly provisioned and released with minimal management effort or service provider interaction.” }[11]. \\
In the NIST special publication on cloud computing, five essential characteristics, three service models, and four deployment models are defined (figure 2) [11] [9].

\begin{figure}[h!]
 \centering
  \includegraphics[scale=0.42]{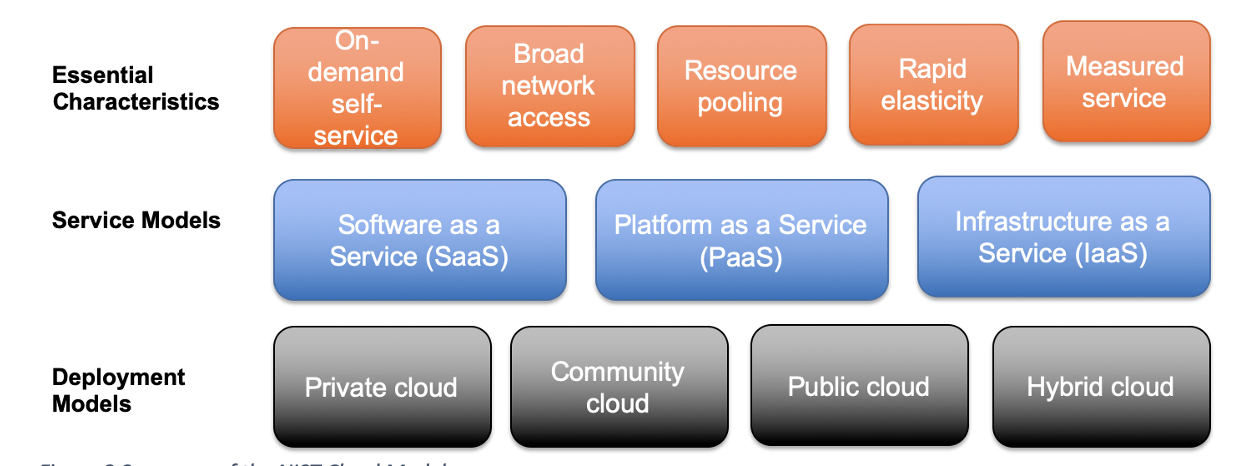}
    \caption{Summary of the NIST Cloud Model}

  \label{fig:CDS}
\end{figure}

\subsection{Essential Characteristics}
The first parts of the NIST cloud model are the Essential Characteristics: on-demand self-service, broad network access, resource pooling, rapid elasticity, and measured service. A cloud implementation should include all of these characteristics.

On-demand self-service is about the ability of a consumer or user to provide the required computing capabilities (such as server time or network storage) by themselves on-demand and automatically without requiring human intervention from the server provider side. 
Next, broad network access means that easy access should be provided through network access by using standard mechanisms. This broad access also refers to the requirement that it should be accessible through a wide variety of different thin or thick client platforms ranging from smartphones and tablets to laptops and workstations.

The characteristic of resources pooling refers to the fact that the provider is running pooled computing resources that can be used to serve multiple customers using a so-called multi-tenant model. The assignment of virtual resources is dynamically executed based on the consumer's requirements. This also means that the user is not knowledgeable about the exact location of the provided resources (e.g., which server or rack is providing the resources). However, because of legal restrictions that might occur, most providers do allow a user to select a designated location on a higher abstraction level to limit the resource location to, for example, a particular country, state, or datacenter. The resources that the cloud provider can provide and dynamically assign are multiple and include storage capacity, CPU and GPU processing, memory, and network bandwidth.

The term rapid elasticity is used to describe the characteristic of the cloud that rapid on-demand scaling of resources is possible within the virtual environment of the consumer. This scaling could be done automatically based on specific triggers when additional resources are needed. This kind of scaling gives a sense of unlimited resources available to the user.

\subsection{Service Models}
The service models describe three different variations of cloud services that provide software, platform or infrastructure through cloud access.

\textbf{Software as a Service (SaaS)}
In short, Software as a Services, or SaaS, deals with running specific applications through a cloud service. This means that a provider provides access to a single application through the cloud service. Al management and control of the underlying infrastructure such as the servers, operating system, storage, etcetera is fully controlled by the provider of the service without any influence on this by the user. A typical example is a web-based e-mail service such as Gmail.

\textbf{Platform as a Service (PaaS)}
Expanding on the SaaS, a Platform as a Service (PaaS), provides a suite of applications, programming languages and other user tools through a cloud service. This means that a user is able to deploy required applications either own or acquired to a cloud service. The underlying cloud infrastructure is still fully controlled by the cloud provider, but the user has influence on what applications should be deployed in the environment.

\textbf{Infrastructure as a Service (IaaS)}
Finally, Infrastructure as a Services (IaaS) provides the most flexible solution by providing access to fundamental computing resources. In this service model the underlying cloud infrastructure is provided by the cloud provider, but the user can control and configure the operating system, storage, and application deployment. Also limited control over selected networking component, for example hosting a firewall, could be granted to the user.

\subsection{Deployment Models}
The Deployment Models proposed by NIST range from Private via Community to Public cloud and allow for Hybrid solutions combining private and public cloud technology. 
The private cloud is the most closed – and assumed most secure – deployment model. In this model, a single organization will be granted exclusive use for multiple consumers belonging to that organization. This does not mean it needs to be owned, managed, and operated by that organization, and all hardware must exist on-premise. Third parties could be involved in one or multiple of these points. 

In the case of a community cloud, the use is not exclusive to one organization but to multiple organizations part of a community. In most cases, such a community will exist of a group of organizations with shared concerns, such as, for example, a group of hospitals in a region or country. Again, ownership, management, and operation can be shared with a third party, and the actual hardware may be on or off-premises.

A public cloud is a cloud infrastructure that is open for use by the general public. This will, in most cases, physically exist on-premise with the cloud provider, which can be a single or combination of several companies, universities, and government organizations.

In a hybrid cloud, the cloud infrastructure is deployed consisting of a combination of cloud models. The different environments are unique entities but can share and exchange data and applications.

\section{AI pipeline}

Because of the complex nature of the process of AI and the different steps involved, research is ongoing to develop more automated and integrated AI pipelines that support the whole process from data collection to deployment [8]. Depending of the situation, each step in such an AI pipeline is prone to automation thus eliminating as many of the manual steps as possible. 
In most cases, there is a distinction between a development and a deployment pipeline. The development pipeline consists of the steps of data collection, data annotation or labelling, data cleansing, training and validation. In a deployment pipeline one or more trained networks will be included to receive (imaging) data and produce a final prediction which again needs to be integrated into the (clinical) database [8].
\subsection{Local implementation}
When implementing AI locally it can be part of the current PACS or EMR environment or run as a private cloud or on-premise service.

\subsubsection{Networking}
Local networks consist of data storage systems, processors, and deployment systems. These components are interdependent, i.e., each system is unable to operate unless it receives a response from other systems. This interdependency makes communication a crucial factor in developing AI pipelines. The ultimate goal of a local network should be to minimize latency and to synchronize the processing among different nodes. If communication in a network is sub-optimal, the network with its costly hardware resources becomes under-utilized and inefficient. 
Several techniques can enhance communication efficiency in a locally implemented network. The first effective technique in high-performance computing is reducing model precision. In many cases, data and model parameters are stored with double precision, which could be converted to a single floating point or less. Since medical data are not required to have double floating-point precision, this conversion does not hurt the model performance. Removing additional integers could help save bandwidth, reduces model size and network burden. [12] 
Another helpful technique to improve data exchange speed is the compression of data transferred between different nodes. This is especially important in (medical) infrastructures where network bandwidths are limited. Some compression techniques utilize lossless methods, i.e., data is fully recoverable after decompression but has limited capacity to save bandwidth usage. Instead, lossy methods are utilized more frequently. One method in deep learning-based AI applications to perform compression is to limit the values of gradient updates to binary values. This method ensures that gradient updates are in the correct direction and transferred data will be as low as one bit. Another technique is to neglect insignificant, small gradient update values and convey only significant gradients. Tao et al. implemented this method for federated systems, applicable in medical AI pipelines [13]. 
An AI pipeline consisting of multiple components requires each component to participate actively, exchange data, and allow access to other components when necessary. Nevertheless, most PACS systems are unable to operate within an active environment. They might, for example, fail when changing IP, hostname, and DICOM attributes since these are hardcoded to work on a pre-defined situation with limited options and minimum flexibility [14].
\subsubsection{GPU and CPU}
Early AI pipelines and deep learning deployments used CPUs to perform computations. AI pipelines were dependent on clusters of CPUs to parallelize computing and improve model efficiency. However, state-of-the-art deep learning pipelines are nowadays relying on GPUs instead of CPUs. For many medical institutes acquiring GPUs is a crucial step in deploying AI pipelines, for both local and cloud-based implementations.
To build an AI pipeline, both CPU and GPU nodes are needed, and tasks are divided between them to reach the optimal point of cost-efficiency. Since GPUs maximize performance, not latency, they have worse latency than CPUs. Hence, GPU nodes are mostly used in the training phase, and CPU nodes are mostly used in the deployment phase when the latency becomes an important factor. Thus, when building large local sites for AI in healthcare, a combination of GPU and CPU devices is preferred. 
\subsubsection{Data Management}
Having high volumes of data requires careful data preparation, storage, processing and exchange between systems when developing local AI pipelines. Data should also be well integrated into existing hospital IT systems in both development and deployment stages.
Developing DL models on existing hospital image archiving systems requires a good integration. This interoperability may be hampered by the co-existence of multiple imaging databases (PACS) in the hospital environment. During development, the DL model can be trained on a data dump of DICOM images to avoid using PACSs in the process. After deployment, it could be needed to ensure all systems within a local/cloud network have the same PACS version, or that PACSs are connected with an intermediary web-based protocol which allows to integrate data from different PACS systems. 
AI systems, similar to clinical experts, require various types of patient data to obtain comprehensive knowledge about a medical situation. For example, in radiology, additional patient EMR data might be required, and AI models cannot be trained properly until they integrate all relevant data. This requires data exchange between different departments within the same hospital. Protocols and data type standardizations should thus be developed for all the relevant departments to facilitate information exchange.\newline\subsubsection{AI Models}
In addition to training datasets, trained AI models should also be stored in the medical IT environment. AI researchers might need future access to the trained models for testing, deployment, enhancement, and migration to a new framework. Thus, models must be accessed, queried, and analyzed efficiently through hospital systems. In larger networks, model interoperability is another concern. Each system might have its own DL framework, which is a burden to develop an AI pipeline. Research is being done to facilitate the interoperability of DL models trained on different frameworks by introducing standardized exchange formats. One popular framework is ONNX [15], which enables various DL frameworks, including Pytorch, Tensorflow, Caffe, MXNET, and CNTK, to share their model properties and co-operate within the same network.

For AI pipelines implemented locally, there is a risk for AI models to be prone to various kinds of bias, including demographic bias (gender, age) or data bias (annotation, equipment, acquisition), thus having limited capacity for generalization. Solutions include using distributed datasets and transfer learning. Current DL models might be designed to work with limited data types (e.g., one modality or annotation protocol). However, with the current pace in imaging techniques, any change has to be expected in model design. Models are required to handle new data formats with minimal interruption. Thus, it is better to avoid static AI models for medical usages and design flexible models.
\subsection{Cloud implementation}
\subsubsection{Development phase}
In the development phase of the AI pipeline, the cloud can be employed to provide compute power (cloud-based learning), data science workspaces, crowd-sourced annotation, or distributed learning.
The process of annotation or labelling of imaging data can be quite time-consuming and difficult to organize because of the low availability of busy healthcare professionals that are needed for expert annotation. Cloud implementations could help in such cases to provide an online annotation environment where experts from all over the world can provide their annotations on the imaging data [7]. Important in such a process is to have a transparent annotation procedure in place which is well documented and uses guidelines to ensure the annotation from different users is uniform [1].

With cloud-based learning, the cloud is used to train the deep learning network by utilizing cloud-provisioned CPU or GPU computation power. The programming and implementation of the deep learning network is done locally and the High-Performance Computing (HPC) environment in the cloud is used to perform the iterative learning process.

Data science workspaces provide access to a full data science environment and allow all development steps to be performed in the cloud. In this case the full programming environment is provided through a virtual machine in the cloud and access to the HPC cloud facilities is also provided.
Because of the move to the cloud it has also become feasible to not limit the learning phase in the development to a single location but to distribute the work over multiple locations and even with different databases at those different locations. This distributed or federated learning is a novel area that could solve issues concerning the limited sharing of data and data privacy, it will be described in more detail separately in the next section.

\subsubsection{Deployment Phase} The deployment phase introduces possibilities such as a single AI algorithm in the cloud, compute power in the cloud, federative deployment of AI and AI platforms [16].

Many AI applications are provided by the vendor as a single AI algorithm in the cloud. This means that the imaging data will be sent to a cloud-based DICOM receiver from the PACS or an upload through a web interface is done. After processing of the data, a report or other generated output (e.g. segmentation files, processed imaging data, etc) will be returned to the user uploading the data. In a 2020 report, Mehrizi et al. showed that of 269 AI applications on the market 32 \% were offered as a cloud- only solution and 46\% offer a choice to be either cloud-based or on-premise [17]. Besides commercial implementations, these kind of cloud-based solutions for online AI applications are also available from research institutions in order to support research. The RECOMIA platform is an example of such a research-based AI application provider that also aims to expand and function as a research marketplace for AI applications developed by other research groups [7]. In the deployment phase, compute power in the cloud could also be used if the required HPC facilities are not available on-premise.

Several federative deployments of AI can be envisioned. One possible solution is the one proposed by the Early Lung Imaging Confederation (ELIC) [6]. They proposed and tested a federative environment where imaging data is distributed over the world at the locations where the data were obtained. A central database is updated on the content of these different systems. Software tools (AI or other) that perform tasks on the decentralized images will be distributed to the relevant nodes in the federative network where automated image analysis is performed. All resulting quantitative measurements are collected on the central HUB where further analysis can be done on the aggregated database.

AI platforms can also provide a centralized environment to manage and run multiple applications that can be accessed by the user through some sort of online marketplace [16]. In such a cloud-based marketplace of AI tools, developers can distribute their tools (after approval by the marketplace holder) to a large customer base without the requirement of their own sales and deployment team [16]. The deployment only needs to be done in the environment of the marketplace, saving the costly and tedious in-hospital connectivity challenges. Also, all workflow and user-interaction is covered through the platform, ensuring that the AI tool developer only needs to focus on the core technology and the API connection to the platform.

For the user, such a platform removes the burden of installing a multitude of different AI applications all performing their own task, application selection and implementation of AI is also simplified to a large extent. The platform will handle the data transfer and ensures the right data is sent to the right application and ensures the data transfer security. The organisation using the platform only has the concern itself with the configuration and implementation of the required data connections once. When connecting to the platform, all AI tools will rely on that single data connection to obtain data. Furthermore, maintenance and future changes will also be limited to that single platform connection (e.g. when migrating to a new PACS or changes in imaging modalities) and updates and upgrades of the AI tools themselves will be handled by the platform provider. The AI tools provided can either be from one specific vendor or from a variety of suppliers including scientific institutions that provide access to research outcome. The platform provider will select AI tools based on pre-defined criteria concerning the legal status of the tools and the quality of the results. A variety of these marketplaces already exist, in their technography study Mehrizi et al. reported 10 of such marketplaces that provide access to other applications in 2019 [17].

\section{Distributed / Federated Learning}
Training deep learning models requires finding optimal values for millions of parameters, and this is time-consuming. Sometimes it is important to distribute the processing on parallel machines, to reduce the training time and improve network efficiency. Besides, there might be privacy considerations and challenges regarding data governance. In such cases, distributed/federated deep learning allows researchers to develop an AI pipeline, without having direct access to data from other institutions. 
\subsection{Parallelization methods}
\textbf{Data parallelization} In data parallelization, each worker (e.g. GPU machine) is given an identical copy of a DL model, and part of the data. To avoid duplication of tasks, data should be split into non-overlapping batches (figure 3). After each model finished its processing, the DL parameters for all workers will be updated. Hence, a strong synchronization among workers is required.

\begin{figure}[h!]
 \centering
  \includegraphics[scale=0.45]{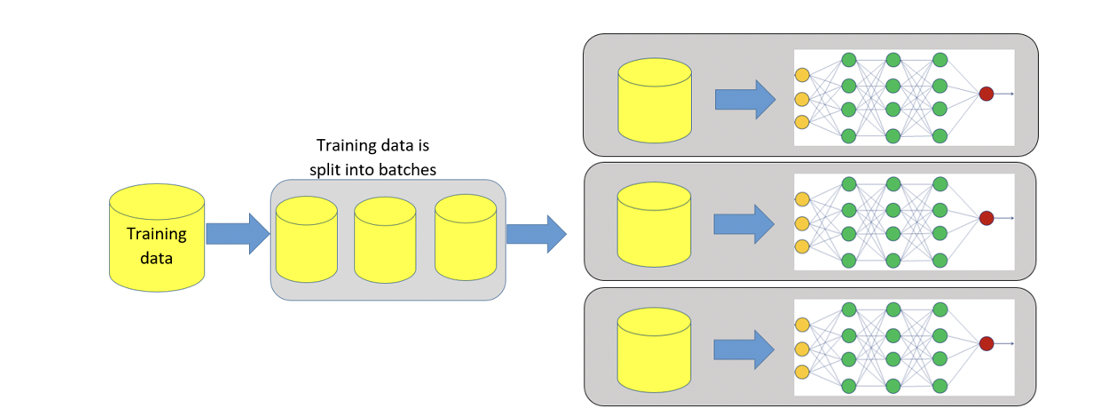}
    \caption{Data parallelism}

  \label{fig:CDS}
\end{figure}
\textbf{Model parallelism} In model parallelism, processors are parallelized to perform operations of a certain part of a deep neural network (figure 4). For example, a neural network can be divided into multiple sections, each processor assigned to one section, and processors exchange information in both forward and backward propagation. A major concern in model parallelism is how to split the network into multiple sections so that the processors work together properly.

\begin{figure}[h!]
 \centering
  \includegraphics[scale=0.40]{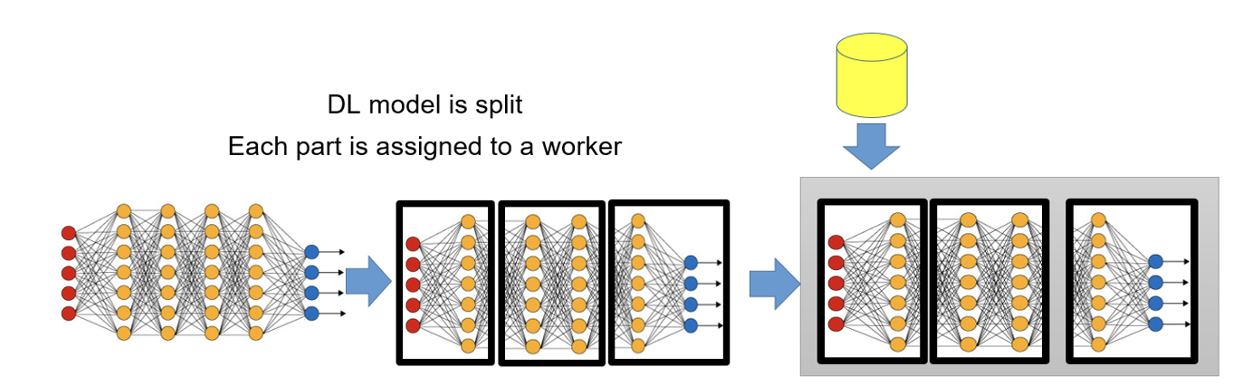}
    \caption{Model parallelism}

  \label{fig:CDS}
\end{figure}

Pipeline parallelism is a combination of model and data parallelism and has the benefits of both methods. In pipeline parallelism, data is divided into non-overlapping mini-batches, and neural network is divided into sections and each section is assigned to a processor (figure 5). In backpropagation, the gradient of a certain mini-batch is propagated through layers and gradient processing is performed on GPUs, associated to those layers.

\begin{figure}[h!]
 \centering
  \includegraphics[scale=0.45]{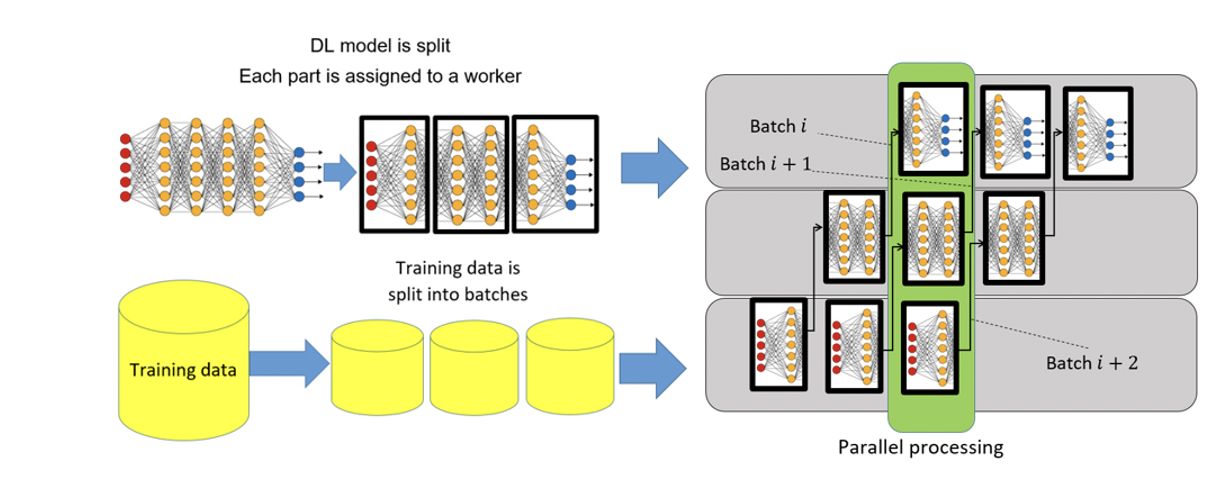}
    \caption{Pipeline parallelism}

  \label{fig:CDS}
\end{figure}

For more complex networks that consist of many layers and complex network architectures, dividing the network or data to distinct parts might not be feasible or reasonable. Thus, large distributed deep learning projects which deal with complex networks, mostly apply Hybrid parallelism - a combination of model, data and pipeline parallelism.

\subsection{Synchronization }
One of the most important questions in distributed deep learning is when to synchronize parameters among workers. Synchronization strategies exert a trade-off between network speed and performance and is challenging.

\textbf{Synchronous training}, workers share their updates after they processed one batch (i.e. after each iteration). This method is widely used in distributed deep learning platforms and has better convergence that other synchronization paradigms. However, synchronous training has the straggler problem; the network speed will be determined by the speed of the slowest worker, and other workers have to wait until the last worker finishes its processing.

\textbf{Bounded synchronous training}, workers train on stale parameters, and parameters resulting from bounded-synchronous training are an approximation of the parameters of synchronous training. However, to prevent that the approximation harms the model accuracy and convergence, the staleness should be bounded. This allows more freedom of the model, and better efficiency, while avoiding the straggler problem. [19]
Another way to update a model is by updating parameters independently or asynchronous, i.e., the model is updated once one worker finishes its batch and sends the gradients to the network, regardless of other workers. This approach might cause problems in convergence. However, the model will be considerably more flexible and training proceeds faster.

\subsection{System architectures }
Having a large number of workers, another major challenge in designing a distributed deep learning architecture is how to synchronize parameters among different workers. Many inter-connected workers exchange considerable amounts of information with each other. The performance of each processors is dependent on the parameters it receives from other processors. Thus, if the system architecture is not designed properly, failure in one worker causes interruption of the whole system. Different system architectures will be discussed in this section.

\textbf{Centralized architectures} are architectures in which every node or worker reports its parameters to a central node, (or nodes), called parameter server (PS). One approach is to divide the model parameters into a few chunks and send parameters belonging to each chunk to a PS. Thus, letting model processors and PSs work in parallel. Notable systems using PS architecture are GeePS [20], DistBelief [21], and Tensorflow [22].

\textbf{Decentralized architectures} work without a parameter server. Processors in decentralized networks, communicate directly with each other (in a fully connected network), or communicate through other processors. In a fully connected network, in which every single worker is connected to all other workers, communication is a major concern. However, alternative topologies exist. For example, a ring topology is implemented in the widely used Horovod [23] framework from Uber.  The major drawback of all topologies other than fully connected topology is that communication between nodes might require involvement of the other nodes and processing time is substantially increased.

Decentralized topologies are much easier to deploy than centralized topologies. There is no need to set-up a PS and complexities of PS planning are avoided. Another advantage is that decentralized topologies are more robust to failure since there is no central hub on which the existence of the network depends. Instead, in decentralized networks, other workers can easily take over other failed workers' duties, and the network goes on without interruption. Decentralized architectures also have disadvantages. Communication is a major problem in decentralized networks. Changing topology from fully connected also introduces other complexities and trade-offs.

Both centralized and decentralized topologies assume that data is available for all workers, and the network data and model parameters are being controlled, either by a central server or by individual nodes. However, in some scenarios, especially in the medical domain, data and model parameters might not be visible to other workers or PS.

\textbf{Federated learning} topology is used in such cases, which is commonly the only available option of distributed deep learning in the medical domains. In federated learning, data is kept locally on each worker and a global network is trained based on the data stored in each server. Each worker reports the updated model parameters based on its dataset to the whole network. This topology is also beneficial for networks with limited bandwidth, because in federated learning, heavy training data is not exchanged in network and only parameters are propagated.

Common topologies of Federated learning are the aggregation server metho, peer-to-peer method, and sequential methods.
In the aggregation server method, a server initializes the model between different workers, each worker computes its gradients and sends them back to the central server. After all of the workers computed their gradients, the model is aggregated and updated from the aggregation server, and the new model is sent back to continue training the next iteration.

\textbf{Peer-to-peer topology} avoids using a server to save the whole model. Instead, a model is initialized from one node, each worker starts training based on its own data. After each node computed its gradients, it reports the calculated parameters to all other nodes. The model is then updated after all of the workers reported their own updates.

\textbf{Sequential methods}, a model is trained on data from one institution and is then adapted to new institutions, with different scanners and protocols. Two forms of sequential methods are domain adaptation and lifelong learning. In domain adaptation, a model is trained on a source domain with rich data. It then will be trained again (fine-tuned) on new samples from the target domain to learn the new distribution. Since the goal of domain adaptation is maximizing performance on the new domain, the model might not be able to preserve its performance on the source domain. Lifelong learning is another variant of sequential methods, aiming to solve this issue. In lifelong learning, a pre-trained model sees a few samples from the new domain, learns the new distribution, while maintains the model’s capability on previous data. This can be done by learning batch normalization parameters for different domains and sharing the convolutional filters [24]. 

\subsection{Discussion on Distributed Learning}
Deep learning models trained on large-scale datasets ensure better clinical performance, generalizability and allows to make less biased decisions. Federated learning helps to provide access to high-quality diagnosis and treatment even in remote locations and the benefit of using data from other institutions. This advantage is significant, especially when dealing with rare diseases. Furthermore, hospitals retain control over their own data, limiting concerns about data leakage or data misuse by third parties although some concerns remain such as what one might learn from the updates to the Parameter Server. For researchers, federated learning helps them to train their models based on a vast amount of diverse data from multiple institutions. This is especially important since even the strongest DL models fail when trained on inadequate data.

However, heterogeneity of data might also have major consequences. For example, the global optimal point might not be the optimal point for each institution, since each institution has a particular group of patients and has some bias in favour of that group. But if an FL framework is trained at multiple institutions, it converges to an optimal point for all parties. And some institutions might not be interested in a global optimal point since they want to serve a specific group. 
Federated Learning training ensures some levels of privacy, but it does not guarantee full security of the model. Some techniques can be applied, but they all add their own complexities to the model and cause trade-offs. Hence, when designing a federated learning paradigm, the desired level of security and performance should always be considered, and an optimal trade-off should be designed based on the level of trust in other institutions, legal limitations, etc.

FL models are also prone to information leakage, even though they do not share direct information from training data. For example, techniques like model inversion, gradient leakage, and adversarial attacks can extract confidential data from the model. If adversaries can track the gradient updates, for example, the values of gradients of a single institute, data from that institute can be stolen.
A high level of safety, reproducibility, and traceability is critical in federated networks. In centralized and local DL models, all the data and systems are in control of a master system. This means that model history, hardware configurations, and hyperparameters are all accessible. As a consequence, faults can be debugged and tracked easily. However, in large scale federated set-ups, not all of the above properties are trackable. Each institution has its own settings, software, network, and infrastructure. One solution to ensure model traceability and integrity is to compel all parties to declare their hyperparameters and training settings.

Finally, in federated topologies, secure data transmission is a necessity. For most strict cases, an additional operator might be required as a ‘trusted broker’ to transmit data between nodes. It exerts an extra cost on the system, which sometimes is undesirable. Encryption of transmitted data is also crucial. Secure data transmission prevents third-parties to access network parameters.
\section{Discussion}
Although very promising in healthcare data exchange, cloud solutions still have to cope with fear and unease about the technology by hospital IT leadership.  On the level of performance this fear is about image latency because of low bandwidth, delay in image access and cloud service disruption by downtime. But also, security issues such as DDoS or hacking attacks and confidential data leakage are major concerns.

That this is still an issue in the current cloud environments of medical imaging was shown by the Health IT Security report in November 2019: about 1.19 billion images were accessible through a very large number of PACS systems that were connected to the cloud and easily accessed by unauthorized users [25].

Security issues prominent in the cloud are secure transfer – including for example data encryption – privacy, confidentiality and integrity of the (imaging) data. The cloud design and implementation should be such that it takes these four points into careful consideration.

Besides security, cloud implementations should also consider safety by providing backup or redundant storage to avoid data loss, high availability with low downtime, and restricted and tracked access by implementation of role-based access and extensive user activity tracking and audit trails.
Although the concerns about security and safety are valid and important, the need to establish imaging biobanks and integrate them with existing biobanks is of utmost importance to facilitate the development of AI-based tools and to increase quantitative, patient centred healthcare. Cloud-based imaging biobanks will help drive AI and deep learning development and the developed tools can be used to further annotate and label other imaging biobanks.

The implementation of data sharing using imaging biobanking and cloud solutions still has to cope with technical, legal and societal challenges. To overcome these, a number of conditions have to be met. One of these is 
the standardization of communication, data format, and lexicons or terminologies which is essential to achieve environments with a high level of interoperability. Interchanging data between cloud solutions and standardization of APIs is still limited [9]. In medical imaging common data standards like DICOM and HL-7 could help in this, together with workflow definitions as provided by the IHE.

AI solutions are a step towards solving many existing problems in the area of healthcare. They provide a strong tool for healthcare experts. Distributed AI solutions, especially federated learning have the capability to provide accurate, safe, privacy-preserving and unbiased models for clinical usage. However, when designing distributed AI pipelines, many considerations should be taken into account. Proper infrastructure, data preparation, and AI model design are necessary for accurate training. How to perform parallelization, and how to enable efficient and secure communication among workers is another challenge which should be addressed.

Other issues to cope with are originating from the legal frameworks as defined by the HIPAA in the United States and the General Data Protection Regulation (GDPR) in the European Union. As an example, the GDPR has a large impact on safety and security of data stored in Imaging Biobanks and Biobanks and requires explicit informed consents of the data subjects, includes the right to be forgotten and requires cross border data transmission and sharing. 
\section{Conclusion}
We need to enable sharing within an institution, between institutions and with patients. To realize this, the cloud plays an important role in fostering better models for fluent exchange of images and information and cloud based medical image sharing and multicentre databases has many advantages. It can be cost effective because of the economy of scale and can improve performance. Furthermore, with online patient health records, it is easier to access and share data between patients and doctors and between doctors. Cloud also offers an increased and more flexible storage capacity. PACS independent storage in the cloud could remove the necessity of tedious data migration projects when moving to another PACS vendor. And finally, consolidating and storing medical image information in single centralized repository in the cloud instead of multiple PACS in different sites means health care providers can quickly access and share images across various departments and organizations resulting in a more patient-centric environment.
\section{References}
\begin{footnotesize}[1]  J. He, S. L. Baxter, J. Xu, J. Xu, X. Zhou and K. Zhang, ``The
practical implementation of artificial intelligence technologies in
medicine,'' \emph{Nat Med,} vol. 25, no. 1, pp. 30-36, 2019. \\\\{[}2{]} Z.
Shi, I. Zhovannik, A. Traverso, F. J. Dankers, T. M. Deist, K. Petros,
M. René, B. Johan, F. Rianne, H. J, A. L, A. Dekker and L. Wee,
``Distributed radiomics as a signature validation study using the
Personal Health Train infrastructure,'' \emph{Scientific Data volume,}
vol. 218, 2019. \\\\{[}3{]} M. A. Levy, J. B. Freymann, J. S. Kirby, A.
Fedorov, F. M. Fennessy, S. A. Eschrich, A. E. Berglund, D. A.
Fenstermacher, Y. Tan, X. Guo, T. L. Casavant, B. J. Brown, T. A. Braun,
A. Dekker and Roelofs, ``Informatics methods to enable sharing of
quantitative imaging research data,'' \emph{Magnetic Resonance Imaging,}
vol. 30, pp. 1249-1256, 2012. \\\\{[}4{]} European Society of Radiology,
``ESR Position Paper on Imaging Biobanks,'' \emph{Insights Imaging,}
vol. 6, no. 4, pp. 403-410, 2015.\\\\{[}5{]} Wilkinson MD, Dumontier M, Aalbersberg IJ, et al. European Society of Radiology,
``The FAIR Guiding Principles for scientific data management and stewardship'' \emph{Sci Data}  2016 Mar 15;3:160018.  2019 Mar 19;6(1):6. 
\\\\{[}6{]} J. L. Mulshine, R. S. Avila, E. Conley, A.
Devraj, L. F. Ambrose, T. Flanagan, C. I. Henschke, F. R. Hirsch, R.
Janz, R. Kakinuma, S. Lam, A. McWilliams, P. M. van Ooijen, M. Oudkerk
and Pastorin, ``The International Association for the Study of Lung
Cancer Early Lung Imaging Confederation,'' \emph{JCO Clinical Cancer
Informatics,} vol. 4, pp. 89-99, 2020. \\\\{[}7{]} E. Trägårdh, P. Borrelli,
R. Kaboteh, T. Gillberg, J. Ulén, O. Enqvist and L. Edenbrandt,
``RECOMIA - a cloud-based platform for artificial intelligence research
in nuclear medicine and radiology,'' \emph{EJNMMI Physics,} vol. 7, no.
51, 2020. \\\\{[}8{]} G. González and C. L. Evans, ``Biomedical Image
Processing with Containers and Deep Learning: An Automated Analysis
Pipeline,'' \emph{BioEssays,} vol. 41, no. 1900004, 2019. \\\\{[}9{]} G. C.
Kagadis, C. Kloukinas, K. Moore, J. Philbin, P. Papadimitroulas, C.
Alexakos, P. G. Nagy, D. Visvikis and W. R. Hendee, ``Cloud computing in
medical imaging,'' \emph{Medical Physics,} vol. 40, no. 7, 2013.
\\\\{[}10{]} T. S. Mathai, Y. Wang and N. Cross, ``Assessing Lesion
Segmentation Bias of Neural Networks on Motion Corrupted Brain MRI,'' in
\emph{MICCAI BrainLes} , 2020. \\\\{[}11{]} P. Mell and T. Grance, ``The
NIST Definition of Cloud Computing,'' \emph{NIST Special Publication ,}
no. 800-145, 2011. \\\\{[}12{]} S. Gupta, A. Agrawal, K. Gopalakrishnan and
Narayanan, ``Deep learning with limited numerical precision.,'' in
\emph{Proceedings of the 32Nd International Conference on International
Conference on Machine Learning}, 2015. \\\\{[}13{]} Z. Tao and Q. Li,
``Communication efficient distributed deep learning on the edge
computing,'' in \emph{USENIX Workshop on Hot Topics in Edge Computing},
Boston , MA, 2018. \\\\{[}14{]} M. Alhajeri, S. Ghulam and S. Shah,
``Limitations in and Solutions for Improving the Functionality of
Picture Archiving and Communication System: an Exploratory Study of PACS
Professionals' Perspectives,'' \emph{Journal of Digital Imaging,} vol.
67, p. 32:54, 2019. \\\\{[}16{]} Blackford Analysis Ltd., ``Adopting a Platform
strategy: Simplify the deployment and management of medical imaging
applications and AI algorithms,'' Blackford Analysis Ltd., Edinburgh,
2019.\\\\{[}17{]} M. H. R. Mehrizi, P. van Ooijen and M. Homan,
``Applications of artificial Intelligence (AI) in diagnostic radiology:
a technography study,'' \emph{European Radiology,} 2020. \\\\{[}18{]} Mayer R, Jacobsen HA. Scalable deep learning on distributed infrastructures: challenges, techniques, and tools. Volume 53 Issue 2021, Article No.: 3pp 1–37,'' \emph{ACM Computing
Surveys,} 2019. \\\\{[}19{]} Cipar J, Ho Q, Kim J, Lee S, Ganger G, Gibson G, Keeton K, Xing E.  ``Solving the straggler problem with
bounded staleness,'' in \emph{14th Workshop on Hot Topics in Operating
Systems}, USENIX Association, Santa Ana, NM, 2013. \\\\{[}20{]} H. Cui, H. Zhang, G. Ganger, P.
Gibbons and E. Xing, ``Scalable deep learning on distributed gpus with a
gpu-specialized parameter server,'' \emph{n Proceedings of the Eleventh
European Conference on Computer Systems,} London,UK, April 2016 Article No.: 4 , 1–16 \\\\{[}21{]}
J. Dean, G. Corrado, R. Monga, K. Chen, M. Devin, M. Mao, A. Senior, P.
Tucker, K. Yang and Q. Le, ``Large scale distributed deep networks,''
\emph{NIPS'12: Proceedings of the 25th International Conference on Neural Information Processing Systems} - Volume 1,December 2012,pp 1223–1231
2012. \\\\{[}22{]} M. Abadi, P. Barham, J. Chen, Z. Chen, A. Davis, J. Dean,
M. Devin, S. Ghemawat, G. Irving and M. Isard, ``Tensorflow: A system
for large-scale machine learning.,'' \emph{: In Proceedings of the 12th USENIX conference on Operating Systems Design and Implementation (OSDI'16).}USENIX Association, USA, 265–283. 2016
\\\\{[}23{]} A. Sergeev and M. Balso, ``Horovod: fast and easy distributed
deep learning in tensorflow,'' \emph{CoRR abs/1802.05799,} 2018.
\\\\{[}24{]} N. Karani, K. Chaitanya, C. Baumgartner and E. Konukoglu, `` A
Lifelong Learning Approach to Brain MR Segmentation Across Scanners and
Protocols,'' in \emph{MICCAI: Medical Image Computing and Computer
Assisted Intervention}, Springer, Cham. ,Granada, Spain , pp 476-484, 2018. \\\\{[}25{]} J. Davis, ``Health IT Security,''
Xtelligent Healthcare Media, 19 November 2019.{[}Online{]}. Available:
https://healthitsecurity.com/news/number-of-exposed-pacs-medical-images-increasing-us-biggest-culprit.
{[}Accessed 27 October 2020{]}.\\\\{[}26{]} ``Data Management Tailored for
Machine Learning Workloads in Kubernetes.,'' Google,{[}Online{]}.
Available: Kubernetes.io.\\\\{[}27{]} I. Chen, F. D. Johansson and D. Sontag
, ``Why Is My Classifier Discriminatory?,'' in \emph{Advances in Neural
Information Processing Systems (NeurIPS)}, 2018.





% \end{thebibliography}

\end{footnotesize}

\end{document}